\documentclass[11pt]{article} 

\usepackage{array}

\usepackage{graphicx}
\usepackage{algorithm,algorithmic}
\usepackage{ijcai17}
\usepackage{times}

\title{What can you do with a rock?\\Affordance extraction via word embeddings}
\author{Nancy Fulda \and Daniel Ricks \and Ben Murdoch \and David Wingate
\\\{nfulda, daniel\_ricks, murdoch, wingated\}@byu.edu
\\
\\Brigham Young University}
\begin{document}
\maketitle

\begin{abstract}
Autonomous agents must often detect \textit{affordances}: the set of behaviors enabled by a situation. Affordance detection is particularly helpful in domains with large action spaces, allowing the agent to prune its search space by avoiding futile behaviors. This paper presents a method for affordance extraction via word embeddings trained on a Wikipedia corpus. The resulting word vectors are treated as a common knowledge database which can be queried using linear algebra. We apply this method to a reinforcement learning agent in a text-only environment and show that affordance-based action selection improves performance most of the time. Our method increases the computational complexity of each learning step but significantly reduces the total number of steps needed. In addition, the agent's action selections begin to resemble those a human would choose.



\end{abstract}

\section{Introduction}

The physical world is filled with constraints. You can open a door, but only if it isn't locked. You can douse a fire, but only if a fire is present. You can throw a rock or drop a rock or even, under certain circumstances, converse with a rock, but you cannot traverse it, enumerate it, or impeach it. The term \textit{affordances}~\cite{gibson77} refers to the subset of possible actions which are feasible in a given situation. Human beings detect these affordances automatically, often subconsciously, but it is not uncommon for autonomous learning agents to attempt impossible or even ridiculous actions, thus wasting effort on futile behaviors.




This paper presents a method for affordance extraction based on the copiously available linguistic information in online corpora. Word embeddings trained using Wikipedia articles are treated as a common sense knowledge base that encodes (among other things) object-specific affordances. Because knowledge is represented as vectors, the knowledge base can be queried using linear algebra. 
This somewhat counterintuitive notion - the idea that words can be manipulated mathematically - creates a theoretical bridge between the frustrating realities of real-world systems and the immense wealth of common sense knowledge implicitly encoded in online corpora. 




We apply our technique to a text-based environment and show that \textit{a priori} knowledge provided by affordance extraction greatly speeds learning. Specifically, we reduce the agent's search space by (a) identifying actions afforded by a given object; and (b) discriminating objects that can be grasped, lifted and manipulated from objects which can merely be observed. Because the agent explores only those actions which `make sense', it is able to discover valuable behaviors more quickly than a comparable agent using a brute force approach. Critically, the affordance agent is demonstrably able to eliminate extraneous actions without (in most cases) discarding beneficial ones.




\section{Related Work}

Our research relies heavily on word2vec \cite{Mikolov2013}, an algorithm that encodes individual words based on the contexts in which they tend to appear. Earlier work has shown that word vectors trained using this method contain intriguing semantic properties, including structured representations of gender and geography~\cite{mikolov2013b,mikolov2013a}. The (by now) archetypal example of such properties is represented by the algebraic expression $vector[$`$king$'$] - vector[$`$man$'$] + vector[$`$woman$'$] = vector[$`$queen$'$]$.



Researchers have leveraged these properties for diverse applications including sentence- and paragraph-level encoding \cite{Kiros2015,Le2014}, image categorization \cite{Frome2013}, bidirectional retrieval \cite{Karpathy2014}, semantic segmentation \cite{Socher2011}, biomedical document retrieval~\cite{Georgios2016}, and the alignment of movie scripts to their corresponding source texts \cite{Zhu2015}. Our work is most similar to \cite{Zhu2014}; however, rather than using a Markov Logic Network to build an explicit knowledge base, we instead rely on the semantic structure implicitly encoded in skip-grams.

Affordance detection, a topic of rising importance in our increasingly technological society, has been attempted and/or accomplished using visual characteristics \cite{song2011,song2015}, haptic data \cite{navarro2012}, visuomotor simulation \cite{schenck2012,schenck2016}, repeated real-world experimentation \cite{Montesano2007,Stoytchev2008}, and knowledge base representations \cite{Zhu2014}.


In 2001 \cite{laird2001} identified text-based adventure games as a step toward general problem solving. The same year at AAAI, Mark DePristo and Robert Zubek unveiled a hybrid system for text-based game play \cite{arkin1998}, which operated on hand-crafted logic trees combined with a secondary sensory system used for goal selection. The handcrafted logic worked well, but goal selection broke down and became cluttered due to the scale of the environment. Perhaps most notably, in 2015 \cite{Narasimhan2015} designed an agent which passed the text output of the game through an LSTM \cite{hochreiter1997} to find a state representation, then used a DQN \cite{mnih2015} to select a Q-valued action. This approach appeared to work well within a small discrete environment with reliable state action pairs, but as the complexity and alphabet of the environment grew, the clarity of Q-values broke down and left them with a negative overall reward. Our work, in contrast, is able to find meaningful state action pairs even in complex environments with many possible actions.





\section{Wikipedia as a Common Sense Knowledge Base}

Google `knowledge base', and you'll get a list of hand-crafted systems, both commercial and academic, with strict constraints on encoding methods. These highly-structured, often node-based solutions are successful at a wide variety of tasks including topic gisting \cite{liu2004}, affordance detection \cite{Zhu2014} and general reasoning \cite{russ2011}. Traditional knowledge bases are human-interpretable, closely tied to high-level human cognitive functions, and able to encode complex relationships compactly and effectively.

It may seem strange, then, to treat Wikipedia as a knowledge base. When compared with curated solutions like ConceptNet~\cite{liu2004}, Cyc~\cite{matuszek2006}, and WordNet~\cite{miller1995}, its contents are largely unstructured, polluted by irrelevant data, and prone to user error. When used as a training corpus for the word2vec algorithm, however, Wikipedia becomes more tractable. The word vectors create a compact representation of the knowledge base and, as observed by~\cite{Bolukbasi2016} and~\cite{Bolukbasi2016a}, can even encode relationships about which a human author is not consciously cognizant. Perhaps most notably, Wikipedia and other online corpora are constantly updated in response to new developments and new human insight; hence, they do not require explicit maintenance.






\begin{figure}\label{genderplot}
\vskip -0.2in
\centering
    \includegraphics[width=\linewidth]{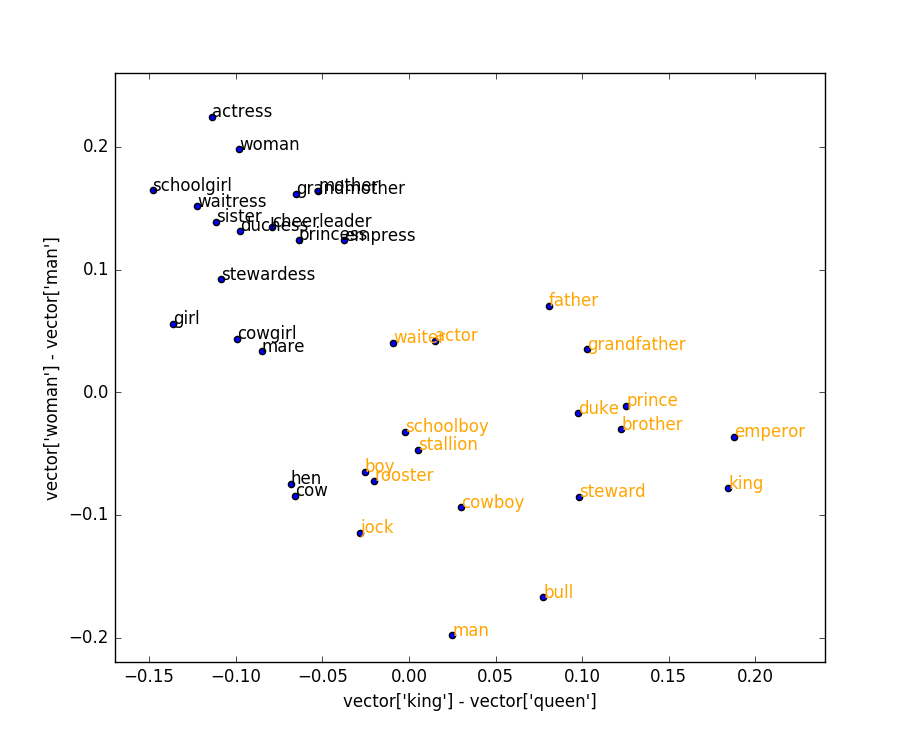}
\caption{Word vectors projected into the space defined by $vector[$`$king$'$] - vector[$`$queen$'$]$ and $vector[$`$woman$'$] - vector[$`$man$'$]$. In this projection, masculine and feminine terms are linearly separable.}
\end{figure}

However: in order to leverage the semantic structure implicitly encoded within Wikipedia, we must be able to interpret the resulting word vectors. Significant semantic relationships are not readily apparent from the raw word vectors or from their PCA reduction. In order to extract useful information, the database must be queried through a mathematical process. For example, in Figure 1 a dot product is used to project gendered terms onto the space defined by $vector[$`$king$'$] - vector[$`$queen$'$]$ and $vector[$`$woman$'$] - vector[$`$man$'$]$. In such a projection, the mathematical relationship between the words is readily apparent. Masculine and feminine terms become linearly separable, making it easy to distinguish instances of each group.


These relationships can be leveraged to detect affordances, and thus reduce the agent's search space. In its most general interpretation, the adjective \textit{affordant} describes the set of actions which are \textbf{physically possible} under given conditions. In the following subsections, however, we use it in the more restricted sense of actions which \textbf{seem reasonable}. For example, it is physically possible to eat a pencil, but it does not `make sense' to do so.

\subsection{Verb/Noun affordances}
So how do you teach an algorithm what `makes sense'? We address this challenge through an example-based query. First we provide a canonical set of verb/noun pairs which illustrate the relationship we desire to extract from the knowledge base. Then we query the database using the analogy format presented by~\cite{Mikolov2013}. Using their terminology, the analogy sing:song::[?]:[x] encodes the following question: If the affordant verb for `song' is `sing', then what is the affordant verb for [x]?

In theory, a single canonical example is sufficient to perform a query. However, experience has shown that results are better when multiple canonical values are averaged.

More formally, let $W$ be the set of all English-language word vectors in our agent's vocabulary. Further, let $N = \{\vec{n}_1, ... , \vec{n}_j\}, N \subset W$ be the set of all nouns in $W$ and let $V = \{\vec{v}_1, ... , \vec{v}_k\}, V \subset W$ be the set of all verbs in $W$.

Let $C = \{ (\vec{v}_1, \vec{n}_1), ... , (\vec{v}_m, \vec{n}_m) \}$ represent a set of canonical verb/noun pairs used by our algorithm. We use $C$ to define an affordance vector $\vec{a} = 1/m \sum_i (\vec{v}_i - \vec{n}_i)$, which can be thought of as the distance and direction within the embedding space which encodes affordant behavior.

In our experiments we used the following verb/noun pairs as our canonical set:

\begin{quote}
[`sing song', `drink water', `read book', `eat food', `wear coat', `drive car', `ride horse', `give gift', `attack enemy', `say word', `open door', `climb tree', `heal wound', `cure disease', `paint picture']
\end{quote}

We describe a verb/noun pair $(\vec{v},\vec{n})$ as affordant to the extent that $\vec{n} + \vec{a} = \vec{v}$. Therefore, a typical knowledge base query would return the $n$ closest verbs $\{\vec{v}_{c1}, ... , \vec{v}_{cn}\}$ to the point $\vec{n} + \vec{a}$


For example, using the canonical set listed above and a set of pre-trained word vectors, a query using $\vec{n} =$ vector[`sword'] returns the following:

\begin{quote}
[`vanquish', `duel', `unsheathe', `wield', `summon', `behead', `battle', `impale', `overpower', `cloak']
\end{quote}

Intuitively, this query process produces verbs which answer the question, `What should you do with an [x]?'. For example, when word vectors are  trained on a Wikipedia corpus with part-of-speech tagging, the five most affordant verbs to the noun `horse' are \{`gallop', `ride', `race', `horse', `outrun'\}, and the top five results for `king' are \{`dethrone', `disobey', `depose', `reign', `abdicate'\}.

The resulting lists are surprisingly logical, especially given the unstructured nature of the Wikipedia corpus from which the vector embeddings were extracted. Subjective examination suggests that affordances extracted using Wikipedia are at least as relevant as those produced by more traditional methods (see Figure 2).



\begin{figure}\label{bentable}
\begin{center}
\begin{tabular}{ | m{3.5em}  m{1.0cm} | m{0.88cm}  m{0.7cm} | m{1.09cm} m{0.9cm} | }
 \hline
 \multicolumn{2}{|l|}{\textbf{Our algorithm}}
  & \multicolumn{2}{|l|}{\textbf{Co-occurrence}} & 
  \multicolumn{2}{|l|}{\textbf{Concept~Net}} \\ \hline \hline
 vanquish & impale    & have  & die    & kill & harm \\ 
 duel &  battle       & make & cut     & parry & fence \\ 
 unsheath & behead    & kill & fight   & strike & thrust \\ 
 summon & wield       & move   & use   & slash & injure \\ 
 overpower & cloak    & destroy & be   & look~cool & cut  \\ \hline
\end{tabular}
\caption{Verb associations for the noun `sword' using three different methods: (1) Affordance detection using word vectors extracted from Wikipedia, as described in this section, (2) Strict co-occurrence counts using a Wikipedia corpus and a co-occurrence window of 9 words, (3) Results generated using ConceptNet's CapableOf relationship.}
\end{center}
\end{figure}

It is worth noting that our algorithm is not resilient to polysemy, and behaves unpredictably when multiple interpretations exist for a given word. For example, the verb `eat' is highly affordant with respect to most food items, but the twelve most salient results for `apple' are \{`apple', `package', `program', `release', `sync', `buy', `outsell', `download', `install', `reinstall', `uninstall', `reboot'\}. In this case, `Apple, the software company' is more strongly represented in the corpus than `apple, the fruit'.

\subsection{Identifying graspable objects}

Finding a verb that matches a given noun is useful. But an autonomous agent is often confronted with more than one object at a time. How should it determine which object to manipulate, or whether any of the objects are manipulable? Pencils, pillows, and coffee mugs are easy to grasp and lift, but the same cannot be said of shadows, boulders, or holograms. 

To identify affordant nouns - i.e. nouns that can be manipulated in a meaningful way - we again utilize analogies based on canonical examples. In this section, we describe a noun as \textit{affordant} to the extent that it can be pushed, pulled, grasped, transported, or transformed. After all, it would not make much sense to lift a sunset or unlock a cliff.



We begin by defining canonical affordance vectors $\vec{a}_x = \vec{n}_{x1} - \vec{n}_{x2}$ and $\vec{a}_y = \vec{n}_{y1} - \vec{n}_{y2}$ for each axis of the affordant vector space. Then, for each object $\vec{o}_i$ under consideration, a pair of projections $\vec{p}_{o_{ix}} = \vec{o}_i$ dot $\vec{a}_x$ and $\vec{p}_{o_{iy}} = \vec{o}_i$ dot $\vec{a}_y$.

The results of such a projection can be seen in Figure 3. This query is distinct from those described in section 3.1 because, instead of using analogies to test the relationships between nouns and verbs, we are instead locating a noun on the spectrum defined by two other words.

In our experiments, we used a single canonical vector, $vector[$`$\mathit{forest}$'$]$ - $vector[$`$tree$'$]$, to distinguish between nouns of different classes. Potentially affordant nouns were projected onto this line of manipulability, with the word whose projection lay closest to `tree' being selected for further experimentation.

Critical to this approach is the insight that canonical word vectors are most effective when they are thought of as exemplars rather than as descriptors. For example, $vector[$`$\mathit{forest}$'$]$ $-$ $vector[$`$tree$'$]$ and $vector[$`$building$'$]$ $-$ $vector[$`$brick$'$]$ function reasonably well as projections for identifying manipulable items. $vector[$`$big$'$]$ $-$ $vector[$`$small$'$]$, on the other hand, is utterly ineffective.






\begin{figure}\label{manipulate}
\vskip -0.2in
\centering
    \includegraphics[width=\linewidth]{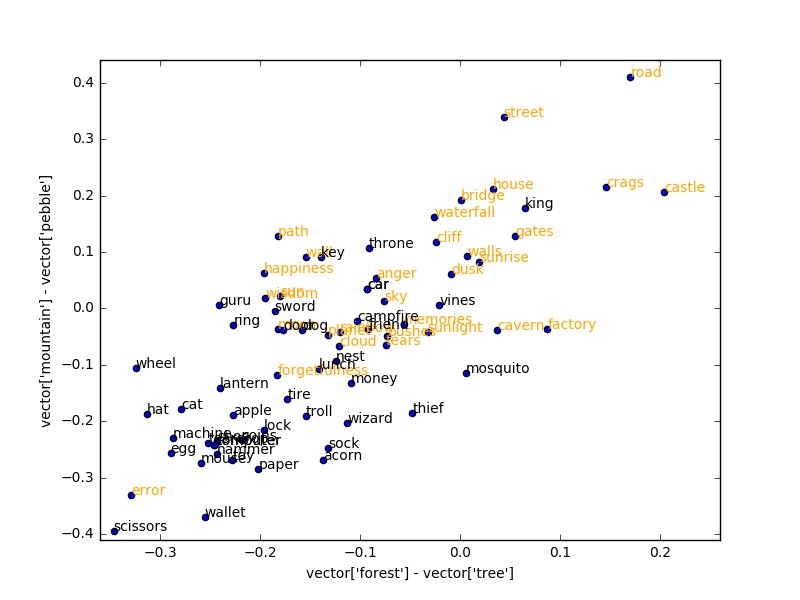}
\caption{Word vectors projected into the space defined by $vector[$`$\mathit{forest}$'$] - vector[$`$tree$'$]$ and $vector[$`$mountain$'$] - vector[$`$pebble$'$]$. Small, manipulable objects appear in the lower-left corner of the graph. Large, abstract, or background objects appear in the upper right. An object's manipulability can be roughly estimated by measuring its location along either of the defining axes.}
\end{figure}

\begin{algorithm}
\caption{Noun Selection With Affordance Detection}
\begin{algorithmic}[1]
\scriptsize 
\STATE $state$ = game response to last command
\STATE $manipulable\_nouns \gets \{\}$
\FOR{each word $w \in state$}
  \IF {$w$ is a noun}
    \IF {$w$ is manipulable}
      \STATE add $w$ to $manipulable\_nouns$
    \ENDIF
  \ENDIF
\ENDFOR
\STATE $noun$ = a randomly selected noun from $manipulable\_nouns$
\end{algorithmic}
\end{algorithm}


\begin{algorithm}
\caption{Verb Selection With Analogy Reduction}
\begin{algorithmic}[1]
\scriptsize 
\STATE $navigation\_verbs$ = [`north', `south', `east', `west', `northeast', `southeast', `southwest', `northwest', `up', `down', `enter']
\STATE $manipulation\_verbs$ = a list of 1000 most common verbs
\STATE $essential\_manipulation\_verbs$ = [`get', `drop', `push', `pull', `open', `close']
\STATE $\mathit{affordant}\_verbs$ = verbs returned by Word2vec that match $noun$
\STATE $\mathit{affordant}\_verbs$ = $\mathit{affordant}\_verbs $ $\cap$ \\ $manipulation\_verbs$ 
\STATE $final\_verbs$ = $navigation\_verbs $ $\cup$  $\mathit{affordant}\_verbs $ $\cup $ \\ $essential\_manipulation\_verbs$
\STATE $verb$ = a randomly selected verb from $final\_verbs$
\end{algorithmic}
\end{algorithm}


\section{Test Environment: A World Made of Words}



In this paper, we test our ideas in the challenging world of text-based adventure gaming.
Text-based adventure games offer an unrestricted, free-form interface: the player is presented with a block of text describing a situation, and must respond with a written phrase. Typical actions include commands such as: `examine wallet', `eat apple', or `light campfire with matches'. The game engine parses this response and produces a new block of text. The resulting interactions, although syntactically simple, provide a fertile research environment for natural language processing and human/computer interaction. Game players must identify objects that are manipulable and apply appropriate actions to those objects in order to make progress.






In these games, the learning agent faces a frustrating dichotomy: its action set must be large enough to accommodate any situation it encounters, and yet each additional action increases the size of its search space. A brute force approach to such scenarios is frequently futile, and yet factorization, function approximation, and other search space reduction techniques bring the risk of data loss. We desire an agent that is able to clearly perceive all its options, and yet applies only that subset which is likely to produce results. 

In other words, we want an agent that explores the game world the same way a human does: by trying only those actions that `make sense'. In the following sections, we show that affordance-based action selection provides a meaningful first step towards this goal. 


\begin{figure}
\centering
    \includegraphics[width=70mm]{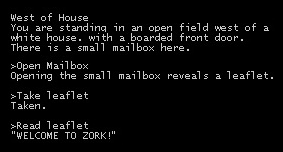}
\caption{Sample text from the adventure game Zork. Player responses follow a single angle bracket.}
\end{figure}






\subsection{Learning algorithm}

\begin{figure*}
\label{final_results}
\vskip -0.2in
\centering
    \includegraphics[width=\linewidth]{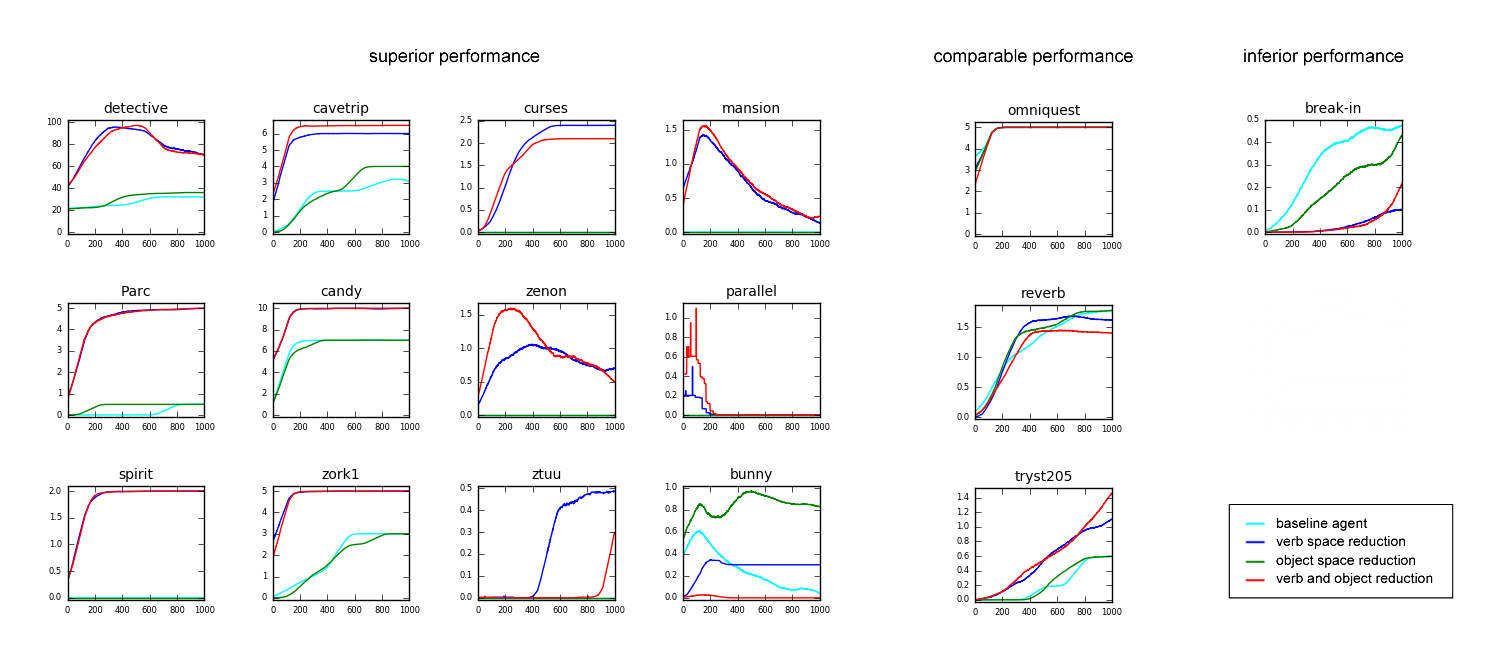}
\caption{Learning trajectories for sixteen Z-machine games. Agents played each game 1000 times, with 1000 game steps during each trial. No agent received any reward on the remaining 32 games. 10 data runs were averaged to create this plot.}
\end{figure*}

Our agent utilizes a variant of Q-learning \cite{watkins1992}, a reinforcement learning algorithm which attempts to maximize expected discounted reward.
Q-values are updated according to the equation
\begin{equation}
\Delta Q(s,a) =  \alpha(R(s,a) + \gamma max_{a}Q(s',a) - Q(s,a))
\end{equation}
where $Q(s,a)$ is the expected reward for performing action $a$ in observed state $s$, $\alpha$ is the learning rate, $\gamma$ is the discount factor, and $s'$ is the new state observation after performing action $a$. Because our test environments are typically deterministic with a high percentage of consumable rewards, we modify this algorithm slightly, setting $\alpha=1$ and constraining Q-value updates such that 
\begin{equation}
Q'(s,a) = max(~Q(s,a),~Q(s,a) + \Delta Q(s,a)~)
\end{equation}
This adaptation encourages the agent to retain behaviors that have produced a reward at least once, even if the reward fails to manifest on subsequent attempts. The goal is to prevent the agent from `unlearning' behaviors that are no longer effective during the current training epoch, but which will be essential in order to score points during the next round of play.

The agent's state representation is encoded as a hash of the text provided by the game engine. Actions are comprised of verb/object pairs:
\begin{equation}
a = v +~`~~\textit{'}~+ o, v \epsilon V, o \epsilon O
\end{equation}
where $V$ is the set of all English-language verbs and $O$ is the set of all English-language nouns. To enable the agent to distinguish between state transitions and merely informational feedback, the agent executes a `look' command every second iteration and assumes that the resulting game text represents its new state. Some games append a summary of actions taken and points earned in response to each `look' command. To prevent this from obfuscating the state space, we stripped all numerals from the game text prior to hashing.

Given that the English language contains at least 20,000 verbs and 100,000 nouns in active use, a naive application of Q-learning is intractable. Some form of action-space reduction must be used. For our baseline comparison, we use an agent with a vocabulary consisting of the 1000 most common verbs in Wikipedia, an 11-word navigation list and a 6-word essential manipulation list as depicted in Algorithm 2. The navigation list contains words which, by convention, are used to navigate through text-based games. The essential manipulation list contains words which, again by convention, are generally applicable to all in-game objects.

The baseline agent does not use a fixed noun vocabulary. Instead, it extracts nouns from the game text using part-of-speech tags. To facilitate game interactions, the baseline agent augments its noun list using adjectives that precede them. For example, if the game text consisted of `You see a red pill and a blue pill', then the agent's noun list for that state would be [`pill', `red pill', `blue pill']. (And its next action is hopefully `swallow red pill').

In Sections 5.1 and 5.2 the baseline agent is contrasted with an agent using affordance extraction to reduce its manipulation list from 1000 verbs to a mere 30 verbs for each state, and to reduce its object list to a maximum of 15 nouns per state. We compare our approach to other search space reduction techniques and show that the \textit{a priori} knowledge provided by affordance extraction enables the agent to achieve results which cannot be paralleled through brute force methods. All agents used epsilon-greedy exploration with a decaying epsilon.

The purpose of our research was to test the value of affordance-based search space reduction. Therefore, we did not add augmentations to address some of the more challenging aspects of text-based adventure games. Specifically, the agent maintained no representation of items carried in inventory or of the game score achieved thus far. The agent was also not given the ability to construct prepositional commands such as `put book on shelf' or `slay dragon with sword'.



\section{Results}
We tested our agent on a suite of 50 text-based adventure games compatible with Infocom's Z-machine. These games represent a wide variety of situations, ranging from business scenarios like `Detective' to complex fictional worlds like `Zork: The Underground Empire'.
Significantly, the games provide little or no information about the agent's goals, or actions that might provide reward. 

During training, the agent interacted with the game engine for 1000 epochs, with 1000 training steps in each epoch. On each game step, the agent received a positive reward corresponding to the change in game score. At the end of each epoch the game was restarted and the game score reset, but the agent retained its learned Q-values.

Our affordance-based search space reduction algorithms enabled the agent to score points on 16/50 games, with a peak performance (expressed as a percentage of maximum game score) of 23.40\% for verb space reduction, 4.33\% for object space reduction, and 31.45\% when both methods were combined. The baseline agent (see Sec.~4.1) scored points on 12/50 games, with a peak performance of 4.45\%. (Peak performance is defined as the maximum score achieved over all epochs, a metric that expresses the agent's ability to comb through the search space and discover areas of high reward.)

Two games experienced termination errors and were excluded from our subsequent analysis; however, our reduction methods outperformed the baseline in both peak performance and average reward in the discarded partial results.

Figures 5 and 7 show the performance of our reduction techniques when compared to the baseline. Affordance-based search space reduction improved overall performance on 12/16 games, and decreased performance on only 1 game.


Examination of the 32 games in which no agent scored points (and which are correspondingly not depicted in Figures 5 and 7) revealed three prevalent failure modes: (1) The game required prepositional commands such as `look at machine' or `give dagger to wizard', (2) The game provided points only after an unusually complex sequence of events, (3) The game required the user to infer the proper term for manipulable objects. (For example, the game might describe `\textbf{something} shiny' at the bottom of a lake, but required the agent to `get shiny \textbf{object}'.) Our test framework was not designed to address these issues, and hence did not score points on those games. A fourth failure mode (4) might be the absence of a game-critical verb within the 1000-word manipulation list. However, this did not occur in our coarse examination of games that failed.

\subsection{Alternate reduction methods}
\begin{figure}
\begin{center}
\begin{tabular}{ | l | l | }
 \hline
\textbf{Affordant selection} & \textbf{Random selection} \\ \hline \hline
decorate glass & continue quantity \\
open window & break sack \\
add table & result window \\
generate quantity & stay table \\
ring window & build table \\
weld glass & end house \\
travel passage & remain quantity \\
climb staircase & discuss glass \\
jump table & passage \\
\hline
\end{tabular}
\caption{Sample exploration actions produced by a Q-learner with and without affordance detection. The random agent used nouns extracted from game text and a verb list comprising the 200 most common verbs in Wikipedia.}
\end{center}
\end{figure}

We compared our affordance-based reduction technique with four other approaches that seemed intuitively applicable to the test domain. Results are shown in Figure 7.

\textbf{Intrinsic rewards}: This approach guides the agent's exploration of the search space by allotting a small reward each time a new state is attained. We call these awards intrinsic because they are tied to the agent's assessment of its progress rather than to external events.

\textbf{Random reduction}: When applying search space reductions one must always ask: `Did improvements result from my specific choice of reduced space, or would \textit{any} reduction be equally effective?' We address this question by randomly selecting 30 manipulation verbs to use during each epoch.

\textbf{ConceptNet reduction}: In this approach we used ConceptNet's CapableOf relation to obtain a list of verbs relevant to the current object. We then reduced the agent's manipulation list to include only words that were also in ConceptNet's word list (effectively taking the intersection of the two lists).

\textbf{Co-occurrence reduction}: In this method, we populated a co-occurrence dictionary using the 1000 most common verbs and 30,000 most common nouns in Wikipedia. The dictionary tracked the number of times each verb/noun pair occurred within a 9-word radius. During game play, the agent's manipulation list was reduced to include only words which exceeded a low threshold (co-occurrences \textgreater~3).

Figure 7 shows the performance of these four algorithms, along with a baseline learner using a 1000-word manipulation list. Affordance-based verb selection improved performance in most games, but the other reduction techniques fell prey to a classic danger: they pruned precisely those actions which were essential to obtain reward.

\subsection{Fixed-length vocabularies vs. Free-form learning}
An interesting question arises from our research. What if, rather than beginning with a 1000-word vocabulary, the agent was free to search the entire English-language verb space?

A traditional learning agent could not do this: the space of possible verbs is too large. However, the Wikipedia knowledge base opens new opportunities. Using the action selection mechanism described in Section 4.1, we allowed the agent to construct its own manipulation list for each state (see Section 3.1). The top 15 responses were unioned with the agent's navigation and essential manipulation lists, with actions selected randomly from that set.

A sampling of the agent's behavior is displayed in Figure
6, along with comparable action selections from the baseline
agent described in Section 4.1. The free-form learner is able
to produce actions that seem, not only reasonable, but also
rather inventive when considered in the context of the game
environment. We believe that further research in this direction
may enable the development of one-shot learning for text-based
adventure games.

\section{Conclusion}

The common sense knowledge implicitly encoded within
Wikipedia opens new opportunities
for autonomous agents. In this paper we have shown that
previously intractable search spaces can be efficiently navigated
when word embeddings are used to identify context-dependent
affordances. In the domain of text-based adventure games, this approach is superior
to several other intuitive methods.

Our initial experiments have
been restricted to text-based environments, but the underlying
principles apply to any domain in which mappings can be
formed between words and objects. Steady advances in object recognition and semantic segmentation, combined
with improved precision in robotic systems, suggests
that our methods are applicable to systems including
self-driving cars, domestic robots, and UAVs.




\begin{figure}
\centering
    \includegraphics[width=\linewidth]{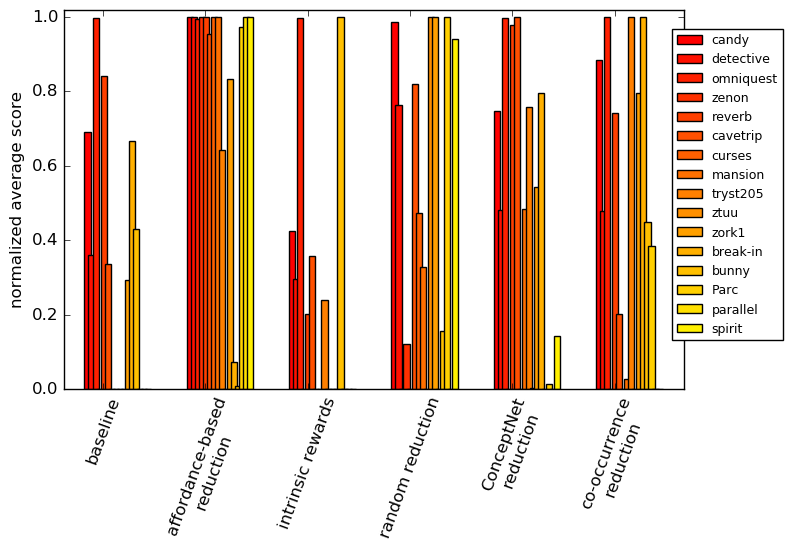}
\caption{Five verb space reduction techniques compared over 100  exploration epochs. Average of 5 data runs. Results were normalized for each game based on the maximum reward achieved by any agent.}
\end{figure}


\section{Acknowledgements}
Our experiments were run using Autoplay:~a learning environment for interactive fiction (https://github.com/-danielricks/autoplay). We thank Nvidia, the Center for Unmanned Aircraft Systems, and Analog Devices, Inc. for their generous support.

\pagebreak
{\small
\bibliographystyle{named}
\bibliography{ijcai17} 

}

\end{document}